THEORETICAL/REVIEW

# Speaker effects in language comprehension: An integrative model of language and speaker processing

**Hanlin Wu**[1] · **Zhenguang G. Cai**[1,2]

**Abstract**
The identity of a speaker influences language comprehension through modulating perception and expectation. This review explores speaker effects and proposes an integrative model of language and speaker processing that integrates distinct mechanistic perspectives. We argue that speaker effects arise from the interplay between bottom-up perception-based processes, driven by acoustic-episodic memory, and top-down expectation-based processes, driven by a speaker model. We show that language and speaker processing are functionally integrated through multi-level probabilistic processing: prior beliefs about a speaker modulate language processing at the phonetic, lexical, and semantic levels, while the unfolding speech and message continuously update the speaker model, refining broad demographic priors into precise individualized representations. Within this framework, we distinguish between speaker-idiosyncrasy effects arising from familiarity with an individual and speaker-demographics effects arising from social group expectations. We discuss how speaker effects serve as indices for assessing language development and social cognition, and we encourage future research to extend these findings to the emerging domain of artificial intelligence (AI) speakers, as AI agents represent a new class of social interlocutors that are transforming the way we engage in communication.

## Introduction

Despite being used in the psycholinguistic literature, the term *speaker effect,* also known as *talker effect*, is often used without being formally defined. It refers to how language comprehension[1] is influenced by the identity of the speaker. For example, when a common name like "Kevin" is mentioned by a colleague, a listener might think of a middle-aged workmate named Kevin, whereas if the same name is mentioned by their school-age son, they are more likely to think of a boy from his class (Barr et al., 2014). Similarly, while it seems natural for a little girl to say she cannot sleep without her teddy bear, hearing the same sentence from an adult man may be unexpected (Van Berkum et al., 2008). These examples illustrate that language is understood in a context that includes the identity of the speaker.

However, using *speaker effect* as an umbrella term often obscures the distinct mechanisms at play in different scenarios. In the "Kevin" example, the effect may arise from the activation of acoustic-episodic memory, linking the name with the voice of a specific speaker (such as the workmate or the school-age son). In contrast, the "teddy bear" example may illustrate the influence of the listener's mental model regarding demographic stereotypes. There is currently a lack of a theoretical framework in which various types of speaker effects can be mechanistically explained and integrated.

To address this issue, we propose a theoretical framework for understanding speaker effects in language comprehension. We begin by noting that the physical basis of speaker effects is the variability in voices across speakers, and that a speaker's voice provides rich information that allows listeners to perceive and identify the speaker. We then consider the interplay between voice and linguistic content by contrasting a *one-system view*, which assumes voice and language processing are integrated, and a *two-system view*, which regards them as independent processes. These two perspectives give rise to two accounts of speaker effects. The *acoustic-episode account*, which aligns with the one-system view, emphasizes the role of acoustic-episodic memory in modulating language comprehension. In contrast, the *speaker-model account*, which aligns with the two-system view, focuses on the influence of the listener's mental model of the speaker on language comprehension. To reconcile these two accounts, we propose an integrative model of language and speaker processing that incorporates the roles of both acoustic-episodic memory and the speaker model. We also illustrate how the speaker model may modulate language comprehension through multiple levels of probabilistic processing. Building on this integrative model, we differentiate between *speaker-idiosyncrasy effects* and *speaker-demographics effects*. The former arise from the listener's familiarity with specific individual speakers, while the latter stem from the listener's accumulated experience interacting with a demographic population. Acoustic-episodic memory and the speaker model contribute to these two types of effects to varying degrees depending on the requirements of different comprehension tasks. Finally, we discuss the potential for using speaker effects as measures for assessing linguistic and socio-cognitive abilities, and suggest extending future research to artificial intelligence (AI) agents as humanlike speakers, as the increasing prevalence of voice-based human-AI interaction may give rise to new types of speaker effects.

[1] In this review, we define *comprehension* as any process that maps lower- to higher-level linguistic representations (Pickering & Garrod, 2013). This includes lower-level processing such as speech perception, as well as higher-level processing like understanding the meaning of a sentence or the speaker's intention.

✉ Hanlin Wu
hanlinwu@cuhk.edu.hk

[1] Department of Linguistics and Modern Languages, The Chinese University of Hong Kong, Sha Tin, N.T., Hong Kong

[2] Brain and Mind Institute, The Chinese University of Hong Kong, Sha Tin, N.T., Hong Kong

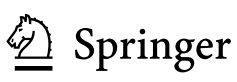

## Voice as the physical basis of speaker effects

Throughout evolutionary history, communication systems in humans and other species have shared a common purpose: to convey the vocalizer's identity and physiological characteristics (Creel & Bregman, 2011). The phenomenon of "speaker" effects is not only prevalent in human communication but is also evident in the animal kingdom. Non-human primates, who share a common ancestor with humans, exhibit vocal recognition systems that enable them to identify individual members within their social groups (Bergman et al., 2003) and perceive cues regarding physical characteristics such as sex, age, and body size (Ey et al., 2007).

Similarly, humans can extract social and biological information from the voice, which is a cornerstone of social cognition (Belin et al., 2004). For example, males with lower-pitched voices are often perceived by other males as more physically and socially dominant, reflecting the importance of voice in male intrasexual competition and mating success (Puts et al., 2006). People rapidly form personality judgments from hearing a new voice, even from just a brief utterance like the word "hello" that lasts less than a second (McAleer et al., 2014). In this section, we explore how human speakers vary in vocal features and how listeners identify speakers on the basis of these features.

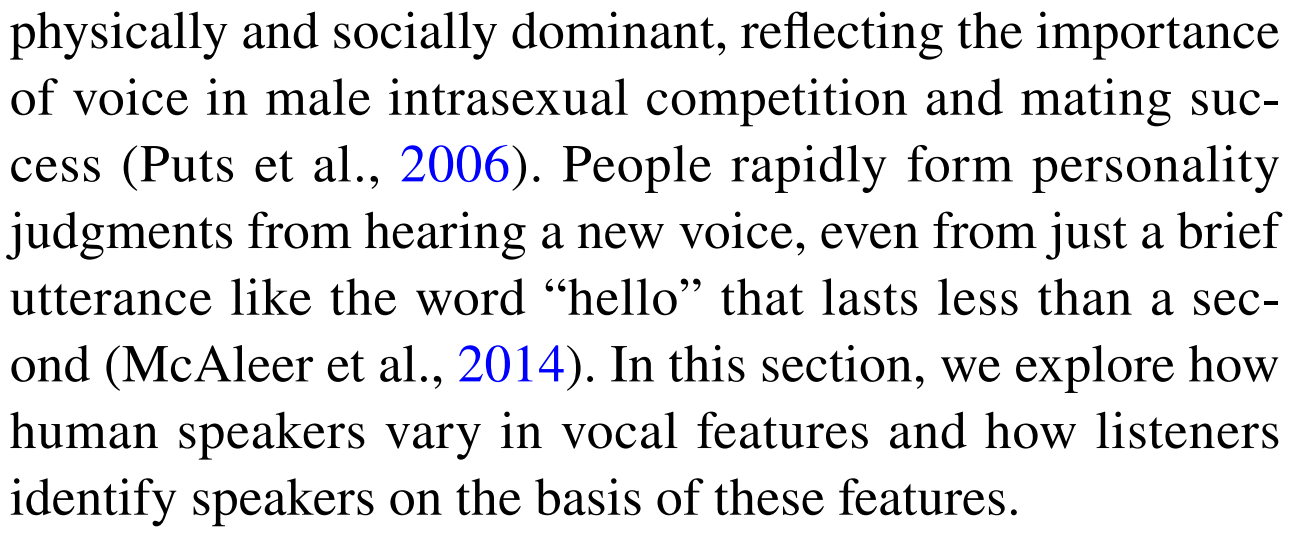

### The source of speaker effects: Speaker variability

In spoken language comprehension, the cognitive processing of a speaker's identity begins with the physical signal. In most cases,[2] speaker effects arise from differences in how acoustic speech signals are produced by different individuals. These differences are known as *speaker variability*, which is often linked to the speaker's unique physiological characteristics and learned behaviors.

Voice production results from interactions between the laryngeal *source* and the vocal tract *filter* (Ghazanfar & Rendall, 2008). Source properties include the vibration of the vocal folds, which determines a speaker's fundamental frequency (f0), a primary cue listeners use to track pitch. Filter properties include the dynamic changes in shape and size of the vocal tract, which acts as a resonator to reinforce certain frequencies. These resonant frequencies of the vocal tract are known as *formants*. The largest acoustic differences in voice among speakers are observed between men and women, and between adults and children (Fitch & Giedd, 1999; Johnson & Sjerps, 2021; Kreiman & Sidtis, 2011). These differences primarily result from physiological factors. For example, men typically have larger larynges with longer and thicker vocal folds compared to women (Hammond et al., 2000), leading to lower rates of vibration and, consequently, lower pitch during speech production (Stevens, 1998). Men also have longer vocal tracts and proportionally longer pharyngeal cavities (Simpson, 2001), resulting in generally lower vowel formant frequencies (Gelfer & Bennett, 2013). Sex-based differences in voice quality, such as breathiness, may arise from variations in subglottal pressure and laryngeal adjustments (Klatt & Klatt, 1990).

However, some differences in voice cannot be explained solely by anatomical differences; instead, they are influenced by social and cultural factors (Munson & Babel, 2019). For example, the acoustic difference between male and female children's speech may be partly attributed to sex-specific articulatory behaviors (Bennett, 1981; Perry et al., 2001). Disparities in fundamental frequencies between males and females throughout their lifespan may arise from culturally ingrained, gender-based pronunciation practices (Whiteside, 2001). A study examining the voice of boys diagnosed with gender identity disorder (now typically referred to as gender dysphoria) showed that they sounded less like control boys, likely due to subtle, learned speech behaviors rather than variations in vocal tract size or vocal cord shape (Munson et al., 2015). Transgender speakers can modify vocal characteristics to align with their identity. For example, transgender women can raise their pitch and adjust resonance despite anatomical constraints, and transgender men can adopt articulation patterns that correlate more with their gender identity (Zimman, 2018). These findings show that the voice can be a social construct mediated by identity rather than strictly a biological outcome.

[2] There are cases where speaker effects arise even in the absence of the speaker's voice. We discuss such cases in *The speaker-model account* section.

Beyond gender differences, the acoustic properties of the human voice change with age, marking significant divergence between adults and children. The development of children's vocalization is primarily due to anatomical maturation of the vocal tract, such as the increase in vocal tract length, which leads to a decrease in formant frequencies. By puberty, notable sex-based differences in vocal tract length emerge, with distinct patterns for males and females (Vorperian et al., 2009). Compared to adults, the speech of children is characterized by elevated pitch and formant frequencies, extended speech segment durations, and increased variability in both timing and frequency spectra (Lee et al., 1999). These distinct patterns enable listeners to identify the age of a speaker upon hearing their voice.

### Speaker identification by voice

Humans can easily identify a person through their voice, and this ability develops very early in life. Human fetuses show increased heart rates in response to their mother's voice and decreased heart rates to a stranger's voice (Kisilevsky et al., 2003). Similarly, newborn babies less than 3 days old can distinguish and show a preference for their mother's voice over that of a female stranger, as indicated by their sucking behavior (DeCasper & Fifer, 1980). For adults, a speaker's voice is a primary acoustic signal that provides rich information about the speaker (Schweinberger et al., 2014). When encountering familiar speakers, listeners can often identify the individual using acoustic cues (Schweinberger et al., 1997). With unfamiliar speakers, listeners can extract demographic attributes from their voice, including their sex (Leung et al., 2018), age (Mulac & Giles, 1996), physical characteristics (Krauss et al., 2002), region of origin (Clopper & Pisoni, 2004b), socio-economic status (Labov, 1973), perceived competence (Rakić et al., 2011; Ko et al., 2009), and sexual orientation (Pierrehumbert et al., 2004).

Listeners can identify a person through their voice remarkably quickly. They show rapid responses differentiating voices from other sounds. In a magnetoencephalography (MEG) study, Capilla et al. (2013) found that listeners begin to show distinct brain responses to vocal and non-vocal sounds as early as 150 ms after stimulus onset. These voice-preferential responses are localized to bilateral mid-superior temporal sulci (mid-STS) and mid-superior temporal gyri (mid-STG), overlapping with the brain regions known as the temporal voice areas. Beyond differentiating voice from non-voice, it takes around 200–300 ms to identify a voice as familiar. An electroencephalogram (EEG) study by Beauchemin et al. (2006) found that familiar voices elicited greater mismatch negativity (MMN) and P3a components, which peaked around 200 ms and 300 ms after stimulus onset, respectively. Furthermore, the categorization of a speaker's social group based on voice also occurs rapidly and interacts with the processing of other vocal cues. Jiang et al. (2020) demonstrated that listeners differentiated vocally expressed confidence as early as approximately 100–200 ms after voice onset for in-group speakers; however, these early differentiation effects were altered or absent for out-group speakers. For newly learned voices, Zäske et al., (2014) found that successful identification was associated with beta-band (16–17 Hz) neural oscillations in central and right temporal regions, starting around 290 ms after stimulus onset. These oscillations appeared to be elicited independently of linguistic content, suggesting a possible dissociation between voice and language processing.

Regarding how voices are represented in memory, some models suggest prototype-based processing as a potential mechanism (e.g., Lavner et al., 2001). In these models, voices are represented in a multidimensional voice space (Petkov & Vuong, 2013). Each dimension represents a vocal feature (e.g., the vocal tract length). The central point of this space represents the prototype voice (Latinus et al., 2013), an average voice formed through prior exposure to different voices. Direct support for this mechanism comes from Lavan et al. 2019b), who found that listeners, after having learned a voice from a specific set of acoustic exemplars, subsequently recognized the untrained mathematical average of that voice better than the specific exemplars they had actually heard. This suggests that listeners automatically construct norm-based prototypes, rather than relying solely on the storage of specific exemplars.

Each voice is represented in the voice space by its deviation from the averaged prototype. Listeners estimate the similarity of an incoming voice to a reference voice based on these deviation patterns (Maguinness et al., 2018). These deviations are compared to stored reference patterns, which may represent a specific speaker (for familiar voices) or broader templates of a demographic group, such as a "young Glaswegian male" (Lavan et al., 2019a).

Furthermore, voice-identity processing is often likened to face-identity processing (Yovel & Belin, 2013), with a person's voice sometimes referred to as their "auditory face" (Belin et al., 2004, 2011; Young et al., 2020). This account, originally developed based on the model of face perception (Bruce & Young, 1986), emphasizes the similarity between voice and face processing (Schirmer, 2018; Young et al., 2020). It suggests that incoming acoustic signals undergo general low-level auditory analysis before being processed in a structural analysis stage where three essential aspects (linguistic, voice, and affective information) are processed through dissociable but interacting pathways. The voice information pathway connects to higher-order semantic nodes of the speaker's identity, which in turn link to other modalities such as the visual system.

### The interplay between voice and linguistic content

Voice is not only a medium for personal identity but also a vehicle for linguistic content (Ladefoged & Broadbent, 1957; Scott, 2019). These dual functions give rise to an interplay between voice and language processing. Over the years, this interplay has been approached from different perspectives, which can be broadly characterized by their focus. Some theories, often grouped as the *two-system view*, suggest that voice and language are processed independently. In contrast, the *one-system view* proposes that voice and language are processed within a single cognitive system from the very beginning. As much of the literature suggests a middle ground where these processes are separable but not wholly independent (e.g., Mullennix & Pisoni, 1990), these two views are perhaps best seen as theoretical endpoints on a continuum.

### The two-system view

Models of voice processing such as the "auditory face" model assume that voice is processed independently from linguistic content. Similarly, abstractionist theories of speech processing (e.g., Liberman & Mattingly, 1985; McQueen et al., 2006) suggest that language comprehension is independent of the processing of paralinguistic information like the speaker's identity. This framework is known as the *two-system view*. In this view, linguistic and paralinguistic (e.g., speaker-related) information are processed in different systems, meaning that indexical information is retained separately (but not completely discarded) (Magnuson & Nusbaum, 2007). To cope with the variability in speech signals from different speakers, the linguistic system engages in *normalization*, an active control process that resolves the many-to-many mapping between acoustics and phonetic categories (Choi et al., 2018; Magnuson et al., 2021) by tuning the speech processing system to speaker-specific acoustic properties (Sjerps et al., 2019).

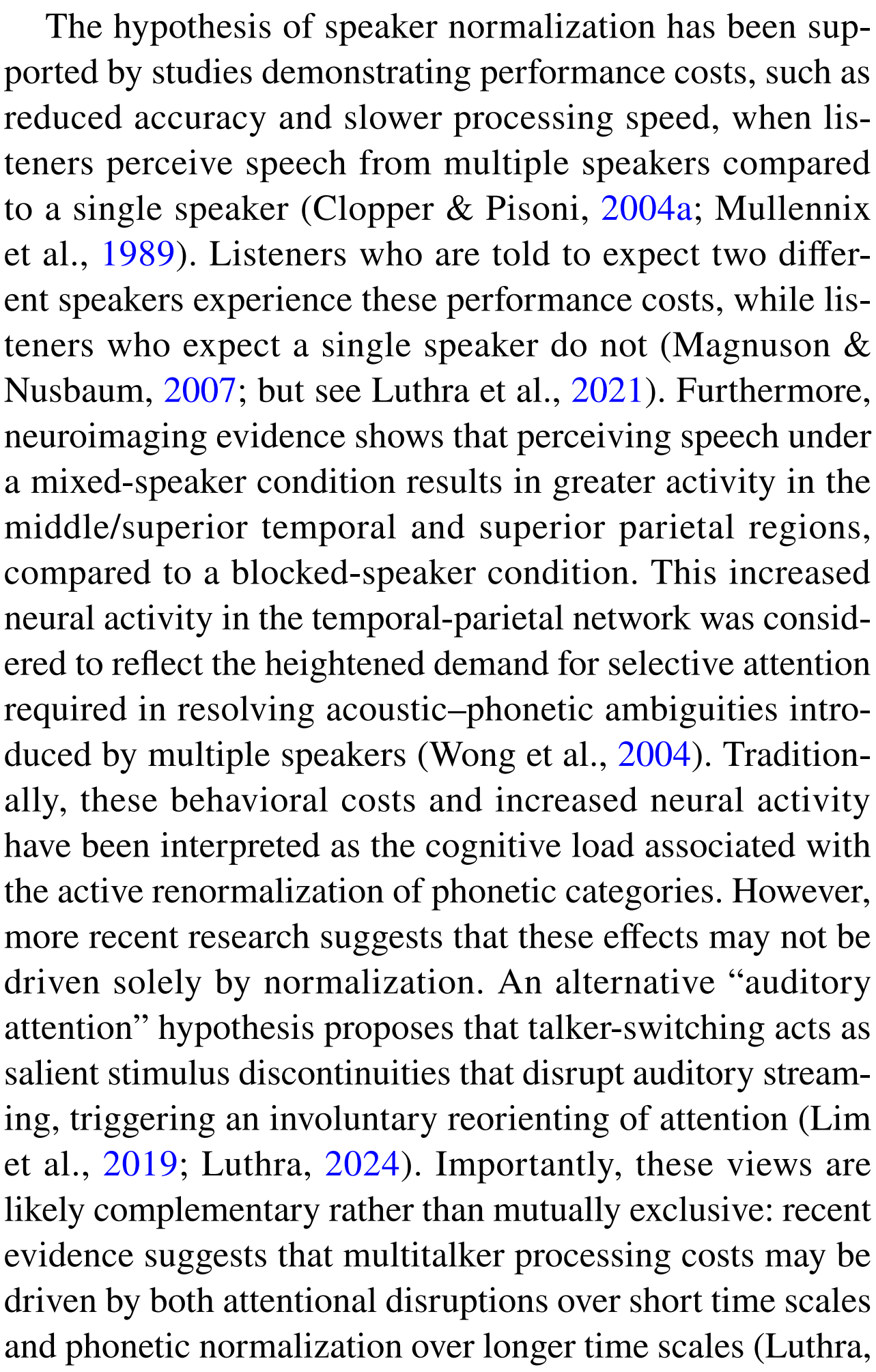

The hypothesis of speaker normalization has been supported by studies demonstrating performance costs, such as reduced accuracy and slower processing speed, when listeners perceive speech from multiple speakers compared to a single speaker (Clopper & Pisoni, 2004a; Mullennix et al., 1989). Listeners who are told to expect two different speakers experience these performance costs, while listeners who expect a single speaker do not (Magnuson & Nusbaum, 2007; but see Luthra et al., 2021). Furthermore, neuroimaging evidence shows that perceiving speech under a mixed-speaker condition results in greater activity in the middle/superior temporal and superior parietal regions, compared to a blocked-speaker condition. This increased neural activity in the temporal-parietal network was considered to reflect the heightened demand for selective attention required in resolving acoustic–phonetic ambiguities introduced by multiple speakers (Wong et al., 2004). Traditionally, these behavioral costs and increased neural activity have been interpreted as the cognitive load associated with the active renormalization of phonetic categories. However, more recent research suggests that these effects may not be driven solely by normalization. An alternative "auditory attention" hypothesis proposes that talker-switching acts as salient stimulus discontinuities that disrupt auditory streaming, triggering an involuntary reorienting of attention (Lim et al., 2019; Luthra, 2024). Importantly, these views are likely complementary rather than mutually exclusive: recent evidence suggests that multitalker processing costs may be driven by both attentional disruptions over short time scales and phonetic normalization over longer time scales (Luthra, 2024).

The two-system view is further supported by neuroimaging evidence indicating that voice and language are processed in separate regions in the brain. The left STG is sensitive to linguistic content, processing phonetic (Yi et al., 2019) and syntactic information (Friederici et al., 2010). This area exhibits flexibility in adapting to different listening environments (Evans et al., 2016). In contrast, the right temporal regions focus more on voice-specific information (Lattner et al., 2005), which is often associated with the speaker's identity. This reflects a hemispheric asymmetry: the left hemisphere is more involved in language processing, while the right is more attuned to nonlinguistic vocal features (González & McLennan, 2007; Schall et al., 2015; Scott, 2019).

Some researchers suggest that this asymmetry arises from differences in the temporal scale of acoustic information processed by the two hemispheres (Creel & Bregman, 2011; Creel & Tumlin, 2011; Poeppel, 2003). The left hemisphere focuses on rapid temporal events, aligning with linguistic elements necessary for speech perception. Conversely, the right hemisphere is sensitive to slower temporal events, which often correspond to the nonlinguistic features that indicate the speaker's identity. However, this purely acoustic account has been challenged by more recent evidence. For example, Myers and Theodore (2017) found that the right hemisphere is recruited to process voice-onset time (a classic rapid temporal cue) when that cue serves as a marker of speaker identity. This suggests that hemispheric specialization might not be driven solely by the physical temporal scale of the stimulus, but also by its functional significance – whether the cue signals linguistic content or vocal identity.

Despite this hemispheric specialization, the two systems must interact to accommodate speaker-specific phonetic variability. Neurobiological evidence shows that the integration of speaker information during speech perception is achieved through coordinated activity of neural networks. Functional connectivities reveal that access to speaker-specific phonetic patterns relies on interactions between the left-lateralized phonetic processing system and the right-lateralized voice processing system, with the right posterior temporal cortex serving as a crucial interface for this integration (Luthra, 2021; Luthra et al., 2023).

It is important to note that although the two-system view suggests separate processing of voice and linguistic content, it does not dismiss the influence of voice on language processing. Instead, it posits that the influence is at most *indirect*, in contrast to the *direct* influence proposed by the one-system view.

## The one-system view

At the other endpoint of the theoretical continuum, the *one-system view* posits that voice and language processing are interdependent and use the same set of representations. According to this view, people learn to distinguish between elements in speech signals that convey meaning and those that identify speakers. Some research also emphasizes that both voice and language processing share an evolutionary root in humans' early ability to recognize individuals from vocal cues (e.g., Creel & Bregman, 2011). The one-system view is represented by exemplar-based theories, which propose that the human memory system, including the mental lexicon, stores detailed records of prior experiences with various stimuli (Medin & Schaffer, 1978; Nosofsky, 1986). When new stimuli are encountered, they are compared with stored exemplars for classification. If a new stimulus matches a stored exemplar, the memory of that exemplar is reinforced; otherwise, a new exemplar is created and stored (Gradoville, 2023).

In a radical version of exemplar-based theories (e.g., Goldinger, 1996, 1998), memory systems (e.g., the mental lexicon) store intact episodes with detailed acoustic traces. Incoming speech signals activate similar acoustic traces in episodic memory, leading to identification. From this standpoint, there is no distinction, as far as speech perception is concerned, in the nature of representations between linguistic units and voice: both are encoded as unified records in the memory system. This comprehensive recordkeeping allows for the emergence of various information clusters, such as words or speakers (Werker & Curtin, 2005).

Listeners direct their attention towards different clusters depending on the task, such as speech perception or voice identification. For example, listeners can flexibly allocate attention to speaker identity or phonemic information depending on the utility of that information for the task at hand (Creel & Tumlin, 2011). This task-dependent processing is also supported by neuroimaging evidence showing that the neural encoding of speech sounds is dynamically reshaped by behavioral goals. Cortical response patterns and the phase of cortical oscillations realign to reflect the specific dimension (speaker vs. vowel) to which the listener is attending (Bonte et al., 2009, 2014).

While radical exemplar-based theories are theoretically appealing, they are often criticized for assuming a memory system that stores vast amounts of information and a comprehension system that requires high computational speed, which may not be economical. Additionally, neurophysiological evidence shows that the brain does encode speech by phonetic categories (Chang et al., 2010) and features such as places and manners of articulation (Mesgarani et al., 2014), as well as specific acoustic features like vowel formants (Oganian et al., 2023). A softer version of exemplar-based theories allows for certain degrees of abstraction (e.g., Ambridge, 2020; Goldinger, 2007; Johnson, 2006), suggesting that both detailed episodic traces and abstract linguistic representations can coexist in the mental lexicon. Nonetheless, the core assumption of exemplar-based theories, and the one-system view in general, is that language processing is *directly* influenced by the acoustic characteristics of the speaker's voice. This is because phonemes and other paralinguistic acoustics are essentially the same and are represented together as one system in the brain.

## Why do speaker effects occur during language comprehension?

The distinct perspectives of the two-system and one-system views give rise to accounts of speaker effects with different theoretical focuses. The two-system view, by separating speaker characteristics from linguistic content, gives rise to a top-down *speaker-model account*, which includes how listeners form phonetic, syntactic, semantic, and pragmatic expectations about the speaker. In contrast, the one-system view, with its focus on holistic memory traces, is most clearly embodied in an *acoustic-episode account*, which highlights the direct episodic influence of the speaker's voice on language comprehension.

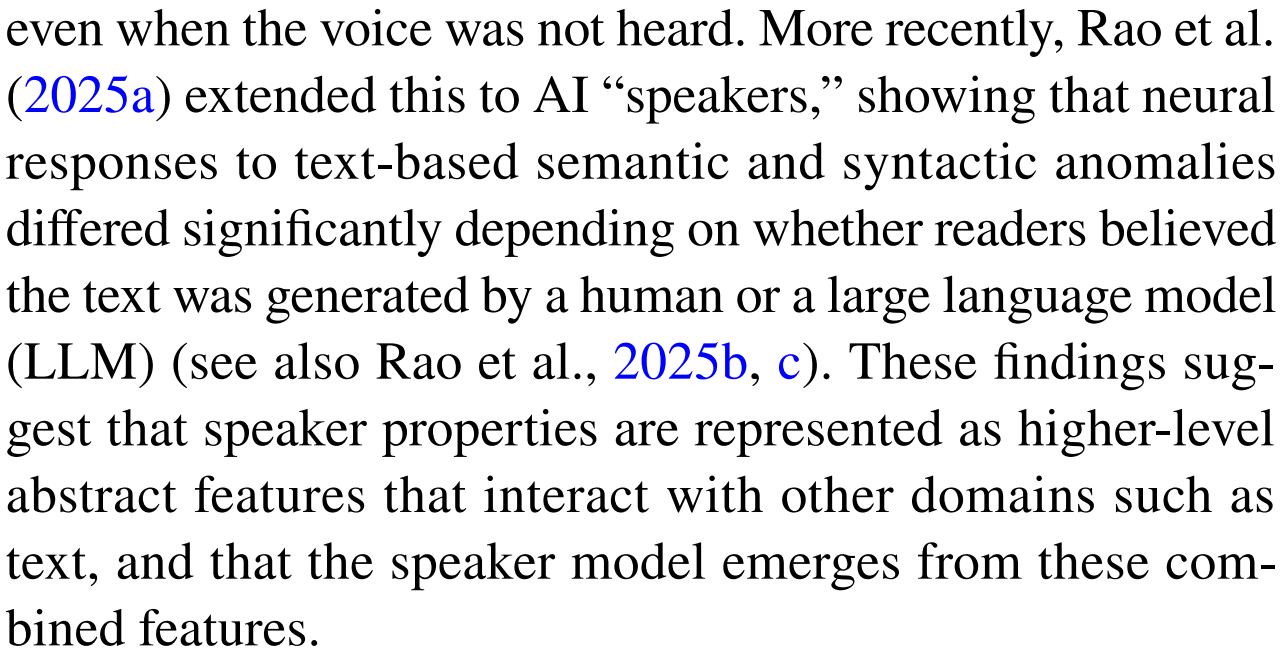

### The speaker-model account

Under the two-system view, the information about a speaker's identity carried by acoustic signals is processed separately from linguistic content. This information enters the voice-processing system and connects to abstract representations related to the speaker, forming a *speaker model*. This model includes the listener's beliefs and knowledge about the speaker, such as their sex, age, socio-economic status, and region of origin. Listeners use this model to form expectations and interpret meaning by integrating the linguistic content with speaker characteristics.

The existence of the speaker model is supported by evidence showing that speaker characteristics can influence language comprehension independently of acoustic variations. For example, Cai et al. (2017) investigated how listeners comprehend cross-dialectally ambiguous English words such as "flat" and "gas." They showed that listeners had more access to the American meaning when these words were spoken by a speaker with an American accent than by one with a British accent. Critically, such speaker effects do not arise from accent details in a word but instead from a mental model listeners have constructed for the speaker (e.g., a British vs. American English speaker): listeners still had more access to the American meaning of word tokens morphed to be accent-neutral as long as they believed the word tokens were produced by an American English speaker (see also Cai, 2022; King & Sumner, 2015).

The speaker model influences comprehension across various modalities, and speaker effects can occur even when acoustic cues are absent. Geiselman and Bellezza (1977) discovered that listeners confused the gender of the speaker with the gender of the agent during a sentence-memorization task. For example, listeners were more likely to remember the speaker being female for the sentence "The queen spent the money" and being male for the sentence "The gentleman entered the house." Fairchild and Papafragou (2018) showed that readers judged under-informative written sentences (e.g., "Some people have noses with two nostrils") as more plausible when they believed the sentences were from a non-native speaker compared to a native speaker. This indicates that expectations about a speaker's linguistic competence modulate pragmatic interpretation even without acoustic input (see also Gibson et al., 2017; Hanulíková et al., 2012). Similarly, Foucart et al. (2019) demonstrated that a brief prior exposure to a speaker's foreign accent modulated the neural processing (N400) of subsequent written sentences attributed to that speaker, suggesting that the reduced reliability associated with the accent was integrated into the speaker model and affected comprehension even when the voice was not heard. More recently, Rao et al. (2025a) extended this to AI "speakers," showing that neural responses to text-based semantic and syntactic anomalies differed significantly depending on whether readers believed the text was generated by a human or a large language model (LLM) (see also Rao et al., 2025b, c). These findings suggest that speaker properties are represented as higher-level abstract features that interact with other domains such as text, and that the speaker model emerges from these combined features.

In addition to these semantic and pragmatic expectations, a speaker model also includes expectations about the speaker's phonetic characteristics. For example, Johnson et al. (1999) demonstrated that participants who were exposed to a gender-neutral voice perceived vowel boundaries differently based on whether they believed the speaker was male or female. This effect occurred when they saw the video clips of a male or female speaker and persisted even when they were simply instructed to imagine a male or female speaker during the task. Similar speaker model effects on speech perception have also been observed regarding a speaker's nationality (Niedzielski, 1999), ethnicity (Staum Casasanto, 2008), and age (Hay et al., 2006). An explanation for this is the "ideal adapter" framework (Kleinschmidt, 2019; Kleinschmidt & Jaeger, 2015). In this framework, listeners solve the "lack of invariance" problem (i.e., the fact that one speaker's acoustic cues for a phoneme, like/s/, differ from another's) by learning a speaker's "generative model." This generative model is a set of statistical distributions for that speaker's phonetic categories. This framework accounts for how listeners recognize a familiar speaker by deploying a stored, speaker-specific generative model, and generalize to a group of similar speakers by using a group-level model (e.g., based on accent or gender) as a starting point for adaptation. The influence of these implicit speaker-phonetic beliefs extends beyond early speech perception; they can also modulate the dynamics of lexical access, such as restricting competition from words that are phonologically incompatible with a speaker's accent (Trude & Brown-Schmidt, 2012), and influence recognition of words (Luthra et al., 2018).

### The acoustic-episode account

Under the one-system view, the intertwined nature of linguistic and speaker representations provides an intuitive explanation for speaker effects. The *acoustic-episode* account, most closely associated with exemplar-based theories, suggests that a speaker's identity influences speech processing by providing a greater or lesser acoustic match to listeners' previous encounters with specific speech episodes (Goldinger, 1996, 1998; Kapnoula & Samuel, 2019; Pufahl & Samuel, 2014). When a word is produced by a familiar speaker, the acoustic details match the listener's episodic memory better than when it is produced by a new speaker (Creel & Tumlin, 2011), leading to speaker effects in speech perception.

In a study by Goldinger (1996), participants were exposed to a list of words spoken by various speakers in a study phase. Later in a test phase, they were presented with another list of words and asked to determine whether each word had been previously heard. The results indicated that they were more accurate in identifying words as previously heard when words were spoken by the same speaker between the study phase and the test phase, compared to when the words were spoken by different speakers. Further research showed that recognition was even better in cases where word tokens were identical (i.e., the same recording), compared to cases where word tokens were not identical (i.e., different recordings) even if uttered by the same speaker (Clapp, Vaughn, Todd et al., 2023b). On the other hand, when learning novel words with similar pronunciations, participants distinguished the words faster when spoken by different speakers during the study phase than by the same speaker (Creel et al., 2008; Creel & Tumlin, 2011); this effect could be detected even when the study phase and the test phase were 24 h apart, suggesting that the speaker's voice may be encoded as part of the mental lexicon (Kapnoula & Samuel, 2019). These findings support the notion that detailed acoustic information, including speaker-specific characteristics, is stored in memory and directly influences speech processing.

Interestingly, the influence of acoustic episodes extends to other acoustic information beyond the speaker's voice, suggesting a highly episodic mechanism in speech perception. In a study by Pufahl and Samuel (2014), participants listened to spoken words accompanied by environmental sounds (e.g., a phone ringing or a dog barking), and made an animacy decision for each word. Later in a test phase, participants' ability to identify acoustically filtered versions of those words was impaired to a similar degree either when the voice changed (e.g., test words were accompanied with the same environmental sound but spoken by a different speaker) or when the environmental sound changed (e.g., test words were spoken by the same speaker but accompanied by a different environmental sound). Similar effects with background noise have been observed for white and sine wave noise (Cooper et al., 2015; Cooper & Bradlow, 2017; Creel et al., 2012; Strori et al., 2018). These findings suggest that lexical and sound representations are deeply integrated, with acoustic-episodic memory directly impacting speech processing.

#### An integrative model of language and speaker processing

The acoustic-episode account and the speaker-model account offer distinct perspectives on the locus and nature of speaker effects (see Creel, 2014 for a similar discussion). The acoustic-episode account assumes that speaker effects arise from bottom-up perceptual processes. In this view, listeners search their memories for the best episodic match to incoming speech signals to determine the word and meaning of a speech token. The speaker's voice, along with other acoustic details, is considered an integral part of the mental representation of spoken words, and these detailed representations directly influence language comprehension. Conversely, the speaker-model account assumes that speaker effects occur in top-down expectation-based processes. According to this account, listeners construct a comprehensive model of the speaker, which includes their beliefs and knowledge about the speaker's characteristics. Listeners then use this model to form expectations and interpret the message by integrating the speaker's characteristics.

While these two accounts may seem contradictory at first glance, they are not mutually exclusive. Speaker effects can take place at multiple representational levels simultaneously (Creel & Tumlin, 2011). Each mechanism can contribute to a speaker effect to varying degrees depending on task requirements. To reconcile these two accounts, we propose an *integrative model of language and speaker processing* that incorporates both bottom-up influences of acoustic episodes and top-down influences of the speaker model on language comprehension.

As illustrated in Fig. 1, incoming sound signals are perceived and form acoustic representations. These acoustic representations are considered unified records of acoustics that do not distinguish between types of information, such as linguistic content or speaker identity. Instead, the acoustic representations capture the complete range of acoustic details present in the speech signal, including both linguistic and paralinguistic information. Listeners can allocate their attention to different aspects of the acoustic representations depending on the context and task requirements, allowing for the emergence of different clusters of acoustics. For example, in a speech perception task, listeners may allocate their attention to distinguishing acoustic clusters between different phonemes and words; in a speaker identification task, listeners may focus on the difference between clusters that represent different speakers.

These acoustic representations proceed through two pathways: one for processing linguistic information and the other for processing speaker information. In the language comprehension pathway, the relevant acoustic features map onto linguistic categories, including smaller units such as phonemes and syllables, and larger units such as words and phrases, ultimately accessing the linguistic meaning. In the speaker perception pathway, the relevant acoustic features map onto representations related to the speaker's characteristics, constructing a model that incorporates information about a specific individual (individual speaker model), or a

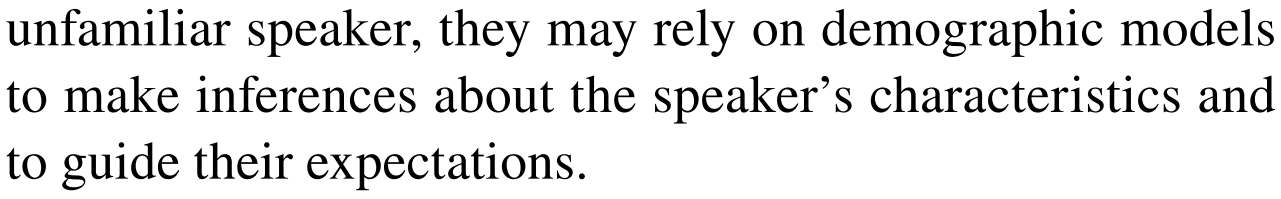

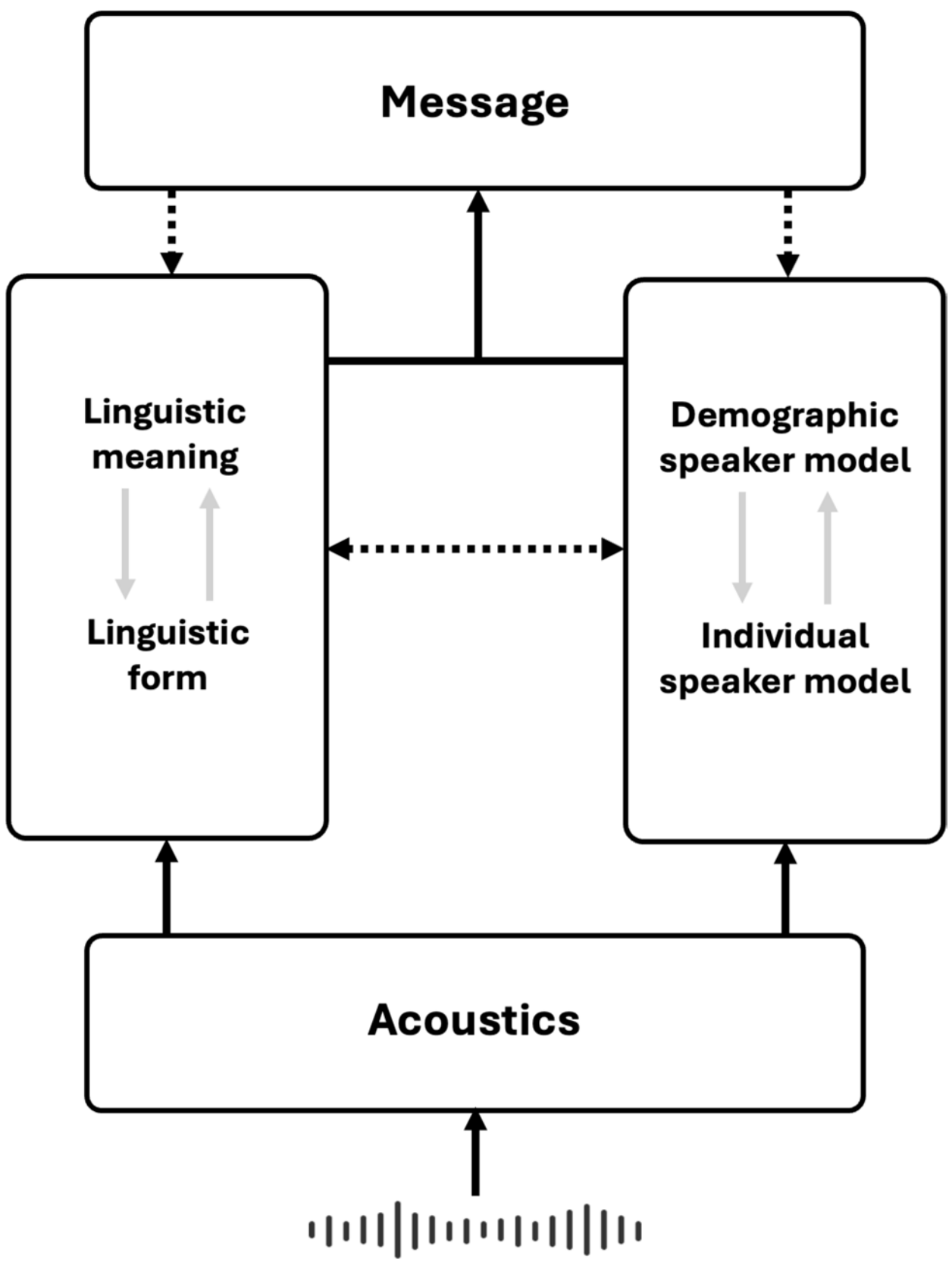

**Fig. 1** Schematic representation of an integrative model of language and speaker processing. Solid arrows indicate the primary feed-forward flow of information involved in constructing the message, moving from acoustic-episodic representations to the formation of the speaker model and linguistic representations. Dashed arrows represent modulatory or feedback influences. Specifically, the speaker model modulates speech perception and meaning access, while linguistic features also inform and modify the speaker model. Additionally, the constructed message can trigger reanalysis of linguistic information and the updating of the speaker model

template model about a social group (demographic speaker model).

An individual speaker model refers to the listener's mental representation of a specific, familiar speaker, encompassing a wide range of information such as the speaker's unique voice characteristics, speaking style, personality traits, background knowledge, and shared experiences with the listener. When a listener encounters a familiar speaker, the acoustic features of the speaker's voice activate the corresponding individual speaker model, which then influences language comprehension by providing a rich context for interpreting the speaker's utterances. On the other hand, a demographic speaker model refers to the listener's mental representation of a social group or category to which a speaker belongs, based on the listener's general knowledge, beliefs, and stereotypes about the characteristics typically associated with members of that group. When a listener encounters an unfamiliar speaker, they may rely on demographic models to make inferences about the speaker's characteristics and to guide their expectations.

Individual and demographic speaker models are not entirely separate; rather, they exist on a continuum and can influence each other. On the one hand, the construction of individual models is usually based on initial demographic models, as a listener's prior experiences with speakers from a particular social group may shape their expectations and biases when encountering a new speaker from that same group; on the other hand, as a listener gains more experience with a particular speaker, they may begin to develop an individual model of that speaker that gradually overrides or modifies the initial demographic model. The relationship between an individual model and a demographic model also aligns with the distinction made in the "ideal adapter" framework between speaker-specific generative models and more general, group-level priors (Kleinschmidt, 2019; Kleinschmidt & Jaeger, 2015).

The speaker model modulates the language comprehension pathway at multiple levels from the top down. At the level of speech perception, the speaker model biases phonetic and lexical processing by applying different prior probabilities to linguistic units. For example, if the speaker model indicates that the speaker might be from a particular dialect region (a demographic model) or is a specific person known to produce/s/with a low-frequency spectrum (an individual model), it may assign higher probabilities to phonetic and lexical variants associated with that speaker (Kleinschmidt, 2019; Kleinschmidt & Jaeger, 2015; Sumner et al., 2014). At the level of meaning access, the speaker model influences meaning interpretation by creating a context that biases dominant word meanings and pragmatic inferences for sentences. For example, if the speaker model suggests that the speaker might be an American English speaker, it may bias the interpretation of ambiguous words or phrases towards meanings more commonly used in American English. Finally, the message is interpreted by integrating the linguistic information with speaker information provided by the speaker model.

It should be noted that the modulation between language and speaker processing is bidirectional. For example, structured phonetic variation in the input also facilitates speaker identification (Ganugapati & Theodore, 2019). Specific linguistic features, such as accent, inform listeners about speaker attributes like region of origin (e.g., identifying a speaker as British vs. American; Cai et al., 2017; Martin et al., 2016), and the speaker model can also be informed by the speaker's lexical and syntactic choices (Porter et al., 2016) and linguistic style (Bradac et al., 1976).

In summary, the proposed model is "integrative" in two senses. In one sense, during language and speaker processing, the bottom-up perception and top-down expectation are integrated, driven by the interaction between detailed acoustic-episodic memory and a more abstract speaker model. In another sense, language and speaker processing are functionally integrated, where the construction of linguistic meaning and the perception of speaker characteristics are not resolved in isolation, but are intertwined throughout comprehension.

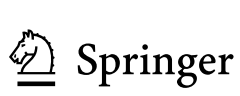

## Probabilistic processing in the integrative model

The integrative model highlights a dynamic, probabilistic interaction between the speaker model and language processing. This dynamic nature can be formalized using a Bayesian framework (see also Kleinschmidt, 2019; Kleinschmidt & Jaeger, 2015), which describes how listeners integrate prior beliefs about a speaker with incoming evidence. This probabilistic processing occurs at multiple levels, including the modulation of speech perception, the modulation of linguistic meaning access, speaker-contextualized message construction, and the updating of the speaker model by the message.

The speaker model modulates speech perception. Formally, the probability of identifying a linguistic form (e.g., a phoneme) given the acoustic input and the perceived speaker identity can be expressed as:

$$p(form|acoustics, speaker) \propto p(acoustics|form, speaker) \times p(form|speaker)$$

Here, the term $p$ (acoustics | form, speaker) is the likelihood of encountering specific acoustic patterns given that a speaker (with a perceived identity) produces a certain linguistic form. The term $p$ (form | speaker) represents the prior probability of the linguistic form given the speaker. This aligns with the "ideal adapter" framework (Kleinschmidt & Jaeger, 2015), where listeners utilize stored statistical distributions associated with that identity to bias lower-level phonetic perception. For example, upon hearing an ambiguous fricative sound, a listener's perception of it as/s/or/ʃ/is based not only on the population-level distribution of the linguistic form but also on the specific phonetic habits of that speaker.

Crucially, this flow of information is not strictly bottom-up but involves a dynamic recalibration process. Listeners use disambiguating lexical information, such as an ambiguous sound (e.g., midway between/s/and/f/) in a context where one interpretation yields a valid word (e.g., "giraffe") and the other a nonword (e.g., "girasse"), to update their beliefs and retune prelexical phonetic categories (Eisner & McQueen, 2005; Norris et al., 2003). This recalibration involves variable spectral cues like fricatives (Kraljic & Samuel, 2005, 2007) and stable temporal cues like stop consonants (Kraljic & Samuel, 2006). This process tracks cumulative input statistics of a speaker's speech over time to iteratively update their phonetic categories (Myers & Mesite, 2014; Tzeng et al., 2021).

The speaker model modulates linguistic meaning access. This involves evaluating the probability of a certain meaning given the linguistic form and the speaker's identity, formalized as:

$$p(meaning|form, speaker) \propto p(form|meaning, speaker) \times p(meaning|speaker)$$

This process is demonstrated in the comprehension of cross-dialectal ambiguous words. Cai et al. (2017) showed that for a word like "bonnet," listeners were more likely to interpret it as a car part (compared to a type of hat) when the speaker was British compared to when the speaker was American. In the current framework, the term $p$ (meaning | speaker) represents the prior probability of the speaker expressing a specific concept. In this case, the prior probability of referring to a car part or a hat may be similar across English speakers (i.e., Americans and British people are equally likely to talk about cars or hats). Consequently, the access to the meaning is largely determined by the likelihood $p$ (form | meaning, speaker). If the speaker is British, the likelihood $p$ (form = "bonnet" | meaning = car part, speaker = British) is high; if the speaker is American, the likelihood shifts: $p$ (form = "bonnet" | meaning = hat, speaker = American) is now high while the likelihood that an American uses "bonnet" for a car part is low (as they would use "hood"). This shift in the likelihood driven by the speaker identity boosts the accessibility of the hat meaning when the listener perceives an American accent.

The listener constructs the final message (i.e., the speaker-contextualized meaning) by integrating the linguistic meaning with the speaker information. This involves a rational evaluation of the joint probability of these two components:

$$p(meaning, speaker)$$

For example, Van Berkum et al. (2008) found that hearing the *meaning* "Every evening I drink some wine" from a child speaker creates a conflict, triggering an N400 effect, because the joint probability $p$ (meaning = drink wine, speaker = child) is low. Wu and Cai (2026) further showed that this joint probability serves as a cue for selecting the appropriate processing strategy. If the joint probability is low but still within a reasonable range (e.g., a social-stereotype violation such as a man talking about himself regularly getting a manicure), the listener engages in effortful integration of social stereotypes and the speaker's identity, reflected as an N400 effect. However, if the joint probability is very low (e.g., a perceived biological violation like a man talking about himself getting pregnant),[3] the listener treats the input as an error and engages in correction/reanalysis, reflected as a P600 effect. As discussed in Wu and Cai (2026), this P600 "error correction" process may itself involve a new probabilistic inference, such as re-evaluating the perceived speaker identity (e.g., misinterpretation of speaker gender based on the voice) or the perceived linguistic content (e.g., misperception of words or inferring a metaphorical interpretation).

[3] It should be noted that while this example is categorized as a biological violation in the literature, it represents an event of very low probability rather than a strict impossibility, as it is suggested that transgender men can become pregnant (Yoshida et al., 2022).

Finally, the speaker model is updated in light of the message. This updating process allows the model to evolve from demographic stereotypes to individualized representations. This can be formalized as a belief update:

$$p(\textit{speaker model}|\textit{message}) \propto p(\textit{message}|\textit{speaker model}) \times p(\textit{speaker model})$$

Here, the posterior belief about the speaker, *p* (speaker model | message), is updated based on the likelihood of observing the current message, *p* (message | speaker model) and the listener's prior beliefs about that speaker, *p* (speaker model). For example, Wu et al. (2025) showed that listeners track the frequency of a speaker making stereotype-incongruent statements. Hearing a child say "I drink whisky every night" for the first time might be a surprising message. However, if the child keeps talking about leading a stereotypically adult lifestyle, the listener updates their speaker model. Wu et al. found that listeners exhibited different neural oscillatory responses depending on the frequency of stereotype-incongruent statements made by a speaker. This indicates that listeners dynamically update their prior speaker model based on the cumulative evidence provided by the message.

### The temporal dynamics of the integrative model

In the integrative model, the interplay between bottom-up acoustic episodes and top-down speaker models occurs rapidly and incrementally as speech unfolds. The influence of acoustic episodes on spoken language processing emerges very early. For example, Creel and Tumlin (2011) demonstrated that listeners use talker-specific acoustic details to distinguish competing words as early as 200 ms after word onset. This suggests that the retrieval of acoustic-episodic traces occurs almost simultaneously with initial phonetic analysis, rapidly constraining lexical selection before the word is fully articulated. This aligns with the general time course of acoustic processing in spoken language comprehension, where acoustic-phonetic analysis occurs within the first 80–200 ms (Tezcan et al., 2023).

The integration of the speaker model with linguistic content does not wait until the end of a sentence; rather, it occurs incrementally as a sentence unfolds. For example, Van Berkum et al. (2008) showed that when a specific word in a sentence mismatches the speaker's identity in terms of social stereotypes, the brain detects this conflict within 200–300 ms of the word's onset, eliciting an N400 effect (similar results were reported in Pélissier & Ferragne, 2022, van den Brink et al., 2012, and Wu & Cai, 2026). This timing suggests that the speaker model is continuously active and integrates dynamically with linguistic content. Although some studies report a P600 effect instead of an N400 in response to speaker-content mismatch, indicating a later stage integration (e.g., Foucart et al., 2015; Lattner & Friederici, 2003), the integrative model interprets these results as a subsequent inference process involving error correction or reanalysis (Wu & Cai, 2026), as discussed in the previous section. Thus, within the integrative model, speaker effects are dynamic: they can manifest as early perceptual biases, concurrent semantic integration, or later error correction, depending on the nature of the input and the listener's rational inference.

## Speaker-idiosyncrasy effects and speaker-demographics effects in language comprehension

When discussing one's *identity*, the term can refer to the idiosyncratic characteristics of an individual speaker, highlighting the unique traits and perspectives that distinguish one person from another. The speaker effects occurring at this level are defined as *speaker-idiosyncrasy effects*. Alternatively, *identity* can refer to the collective attributes of a demographic group, reflecting shared characteristics typical of a specific social, ethnic, gender, or age group. Speaker effects at this level are defined as *speaker-demographics effects*.

However, it is important to note that this distinction is not binary. As similarly proposed by Kleinschmidt (2019), speakers can be conceptualized within a hierarchy of group membership. Listeners' beliefs about a speaker become more specific as the grouping becomes more precise, moving along a continuum from broad demographic categories (e.g., gender, ethinicity) to highly specific idiosyncratic traits. In this view, demographic representations emerge from the experience of interacting with individuals within a demographic group, while these demographic features can, in turn, serve as a basis or prior for forming expectations about a specific individual. In this section, we review studies that examine speaker-idiosyncrasy effects and those that explore speaker-demographics effects.

### Speaker-idiosyncrasy effects

The speaker-idiosyncrasy effect refers to how a speaker's unique characteristics, along with the listener's prior experience with that speaker, can influence language comprehension. Research shows that speech is more intelligible from a familiar speaker than from an unfamiliar speaker, a phenomenon known as *familiar talker advantage* (Domingo et al., 2020; Souza et al., 2013). Evidence suggests this advantage relies on precise, linguistically specific knowledge, as listeners trained on a voice in one language did not show improved intelligibility for that speaker in another language (Levi et al., 2011). Furthermore, this advantage appears not to require explicit recognition of the speaker's identity, as listeners retained the intelligibility advantage even when acoustic manipulations (e.g., of vocal tract length) prevented them from consciously identifying the voice (Holmes et al. 2018). The advantage was also found to be context-dependent, manifesting most strongly when a competing masker is linguistically similar (e.g., speech) rather than dissimilar (e.g., noise), and correlates with the listener's learning accuracy of the voice (Levi et al., 2019). Neuroimaging evidence further shows that familiar voices elicit more robust neural representations in the posterior STG and MTG (Holmes & Johnsrude, 2021), regions known to represent phonetic categories. These results highlight the influence of both acoustic-episodic memory and the speaker model. The fact that the intelligibility benefit survives without explicit speaker identification (Holmes et al., 2018) suggests that low-level acoustic-episodic traces can directly facilitate processing. However, the linguistic specificity of the effect (Levi et al., 2011) indicates that the speaker model must also provide precise, speaker-specific priors for phonetic categories to resolve ambiguity.

Beyond speech intelligibility, research shows that word recognition is faster and more accurate when words are spoken by the same speaker during both learning and test. Craik and Kirsner (1974) had participants listen to a string of words and decide if each word had appeared earlier in the sequence (i.e., whether the word was repeated). Repeated words were spoken either by the same speaker or by a different speaker. They found that participants' responses to the words repeated by the same speaker were more accurate than responses to those repeated by a different speaker. This result was replicated in further studies (Clapp, Vaughn, Sumner et al., 2023a, Clapp, Vaughn, Todd et al., 2023b; Goh, 2005; Goldinger, 1996; Palmeri et al., 1993). In this case, the acoustic match between the initial and repeated tokens of the same word is better when spoken by the same speaker than by different speakers, which leads to more efficient word recognition.

However, the influence of these speaker-specific acoustics on word recognition is not unconditional. Research suggests that these effects are often time-dependent, emerging primarily when processing is relatively slow or difficult. For example, McLennan and Luce (2005) found that these effects emerged in slow responses but were reduced in fast ones. This suggests that abstract phonological representations dominate early, rapid processing (e.g., quickly identifying a clearly spoken word based solely on its phonemes), while specific indexical details emerge only during later, slower processing (e.g., relying on memory of a specific speaker's voice to help identify a difficult word). This idea was further supported by evidence showing that speaker effects were stronger for the speech of dysarthric individuals than for control individuals, as the increased processing time required to decode degraded speech signals enhanced the retrieval of detailed acoustic-episodic traces (Mattys & Liss, 2008). Theodore et al. (2015) further showed that even when processing is fast, speaker effects emerge if listeners explicitly attend to speaker details during encoding (e.g., actively focusing on *who* is speaking rather than just *what* is being said).

Speaker-idiosyncrasy effects also occur in higher-level comprehension tasks, such as referent label processing. Typically, listeners expect speakers to consistently use the same label when referring to the same object (Brennan & Clark, 1996; Shintel & Keysar, 2007). For example, if a speaker initially refers to a piece of furniture as a "couch," listeners anticipate that the speaker will continue using this label, rather than switching to an alternative label like "sofa." When speakers occasionally switch to a different label, comprehension can be disrupted (Barr & Keysar, 2002). Experiments in referent label processing usually involve two phases: initially, a speaker uses a label for an object; later, either the same or a different speaker uses the same or an alternative label for that object – a process known as *label switching*. Evidence shows that the disruptive effect of label switching is modulated by the speaker's identity. Metzing and Brennan (2003) found that when hearing a new referent label, listeners were slower to find the object when the new label was uttered by the original speaker than by a different speaker. This finding has been replicated in further studies using behavioral (Brown-Schmidt, 2009; Horton & Slaten, 2012; Kronmüller & Barr, 2007, 2015) and neurophysiological measures (Bögels et al., 2015). In the context of the integrative model, these studies suggest that listeners develop an individual speaker model based on their experience with a specific speaker's language use. This model encompasses information about the speaker's prior label usage. When the same speaker switches to a new label, it violates the listener's expectations based on their mental model of that speaker, leading to a disruption in comprehension. In contrast, when a different speaker uses a new label, the listener may not have a well-established model for that speaker, resulting in less disruption.

Another example where individual speaker models influence language comprehension is *perspective modeling* (also known as *perspective taking*). Listeners actively consider what the speaker can physically see when interpreting messages. In a scenario described by Brown-Schmidt et al. (2015), a speaker and a listener sit at a table on which there are two red triangles and one blue triangle. One of the red triangles is blocked from the speaker's view but is visible to the listener. When the speaker instructs the listener to move the "red one," it would not be ambiguous if the listener models the speaker's perspective, considering what the speaker can see. Studies have shown that perspective modeling significantly affects language comprehension, especially in referent disambiguation (Brown-Schmidt, 2012; Brown-Schmidt et al., 2008; Hanna et al., 2003). This perspective modeling can be considered part of an individual speaker model that encompasses the listener's understanding of what the speaker knows and does not know (Clark, 1996; Heller et al., 2012; Wu & Keysar, 2007). This aspect of the individual speaker model is constructed through the listener's experience with the specific speaker and their shared context, and likely shares cognitive mechanisms with the spontaneous modeling of co-listeners (Jouravlev et al., 2019; Rueschemeyer et al., 2015). When the listener encounters an ambiguous referent, they can use their individual speaker model to infer the speaker's intended meaning based on their knowledge of the speaker's perspective.

In a proper name comprehension study by Barr et al. (2014), pairs of friends played a communication game in which one friend (addressee) identified a target person from four photos based on a name spoken by their friend or a stranger. The addressee was informed whether the name was chosen by their friend or the stranger. Results showed that addressees identified the target more quickly when the name was spoken by their friend, possibly due to a better match of acoustic details, as they were more familiar with their friend's voice. Meanwhile, responses were slower when told the name was not chosen by the speaker but by the other person (e.g., a friend speaking a name from a stranger), reflecting the addressee's effort to verify the speaker model regarding whether the speaker knows the target person or not. In this case, the listener's mental model of their friend includes knowledge about the friend's social network and familiarity with specific individuals. When the friend speaks a name chosen by the stranger, it conflicts with the listener's mental model of the friend, prompting them to engage in additional processing to verify the speaker's knowledge of the target person.

### Speaker-demographics effects

The speaker-demographics effect refers to how language comprehension is influenced by the collective attributes of a group of speakers who share characteristics typical of a specific social, ethnic, gender, or age population. In the referent label processing studies discussed in the previous section, researchers manipulated the speaker's identity by contrasting whether the speaker who switched referent labels was the one who established the original label in the first place. This involves comparing specific individuals within the same demographic group (e.g., adult speaker A vs. adult speaker B). In contrast, studying speaker-demographics effects involves comparing speakers from different demographic backgrounds (e.g., an adult vs. a child) to examine how group-level expectations influence processing.

Wu et al. (2024) used event-related potentials (ERPs) to explore whether listeners expect a child speaker to be less likely to switch labels compared to an adult speaker, based on the common belief that children are less flexible in language use. They used pictures with alternative labels (e.g., a piece of furniture can be labeled either as a "couch" or as a "sofa"). Each picture was shown twice across two phases. In the establishment phase, participants heard either an adult or a child label a picture and judged whether the label matched the picture. In the test phase, the same speaker either repeated the original label or switched to an alternative label, and participants again judged the label's match to the picture. ERP results showed that switched labels elicited an N400 effect compared to repeated labels. Importantly, the N400 effect was larger with a child speaker than with an adult speaker, indicating greater difficulty in comprehending switched labels from children than from adults.

In this case, although the possibility that the acoustic difference between the original label and the alternative label might be larger for a child speaker than for an adult speaker cannot be entirely ruled out, the speaker-demographics effect here is likely driven primarily by listeners' modeling of the speaker's linguistic flexibility. Specifically, it can be attributed to the listener's demographic speaker model, which incorporates general beliefs and expectations about the linguistic flexibility of different age groups. When listeners encounter a child speaker, their demographic model suggests that children are less likely to switch labels compared to adults. This top-down influence of the demographic speaker model leads to greater processing difficulty when a child speaker violates this expectation by switching labels.

ERP studies also show that the speaker demographics modulate sentence comprehension. In an early study, Lattner and Friederici (2003) asked participants to listen to self-referential sentences that expressed a stereotypically gendered idea, including stereotypically masculine sentences such as "I like to play *soccer*" or stereotypically feminine ones such as "I like to wear *lipstick*." Each sentence was spoken by both male and female speakers. They found that the mismatch between the speaker's biological sex (as inferred from their voice) and the stereotypically gendered sentence elicited a P600 effect at the critical words at the end of sentences (e.g., "soccer" spoken by a female speaker and "lipstick" spoken by a male speaker). Van Berkum et al. (2008) used a similar paradigm and tested more demographic attributes, including age and social status. They contrasted sentences such as "Every evening I drink some *wine* before I go to sleep" spoken by an adult speaker versus by a child speaker. Their results showed that the mismatch between speaker demographics and the linguistic content elicited an N400 effect at the critical word "wine," similar to the classic N400 effects elicited by semantic anomalies (Kutas & Hillyard, 1980; Van Berkum et al., 1999) and world knowledge violations (Hagoort et al., 2004). These speaker-demographics effects on sentence comprehension have been replicated by further studies using similar paradigms (Foucart et al., 2015; Martin et al., 2016; Pélissier & Ferragne, 2022; Tesink, Petersson et al., 2009b; van den Brink et al., 2012; Wu & Cai, 2026).

These findings can be explained by considering the role of the demographic speaker model in sentence comprehension. As the sentence unfolds, listeners incrementally integrate the sentence meaning with their knowledge about the speaker's demographic background, which is captured by the demographic speaker model. When the critical word in the sentence conflicts with the expectations generated by the demographic model (e.g., a child speaker talking about drinking wine), it elicits an N400 effect. This effect has been interpreted by most authors as reflecting increased difficulty in integrating the unexpected word into the current speaker context. However, others might interpret the N400 as a lexico-semantic prediction error (DeLong et al., 2005; Nour Eddine et al., 2024). In this view, the speaker model generates probabilistic predictions about likely upcoming words, and the N400 amplitude indexes the degree of mismatch or "surprise" arising when the bottom-up input conflicts with these top-down predictions.

Another example of such population-specific word frequency effects is demonstrated in the study by Walker and Hay (2011), in which participants completed an auditory lexical decision task where they listened to words that were more prevalent among older people (e.g., "knitting") and words that were more prevalent among younger people (e.g., "lifestyle"). All words were presented in the voices of both older and younger speakers. They found that participants responded faster and more accurately when the age of the voice matched the typical age of the word (see Kim, 2016, for a similar finding). The authors interpreted this finding within an exemplar framework, suggesting that lexical access is facilitated when the specific phonetic detail of the input matches the generalized acoustic detail of the listener's stored exemplars. They argued against a top-down semantic priming account (e.g., the speaker model) by demonstrating that the effect was predicted by objective corpus frequency ratios but not by explicit post hoc ratings of "word age." However, without specific information about the time course of the effect, it remains possible that the speaker model exerts an implicit top-down influence not captured by explicit ratings.

## Speaker effects as indices of language ability and socio-cognitive traits

Despite not directly focusing on the underlying mechanisms of speaker effects, some studies utilize speaker effects as indices for assessing other cognitive abilities. One such ability is language ability, where the influence of acoustic details can reflect the development of an individual's mental lexicon. Another is socio-cognitive ability, which is often linked to the robustness of the speaker model during communication.

### Acoustic-detail effects in phonetic learning

An essential component in language acquisition is learning what elements of speech signals (e.g., phonemes) differentiate meanings. Theoretically, a fully abstract linguistic system would normalize variability that does not distinguish one linguistic unit from another. The presence and magnitude of speaker effects, especially sensitivity to acoustic details during speech perception, can indicate whether language learners have achieved linguistic abstraction. It also reflects whether they can efficiently process linguistically relevant information without being overly influenced by extralinguistic factors like speaker variability. In this sense, attenuated speaker effects (e.g., less disruption caused by speaker changes) may indicate more successful generalization.

During the initial stages of language acquisition, spoken word representations are highly acoustic. This makes it challenging for infants, the primary language learners, to generalize beyond specific acoustic details of their language input. To assess infants' ability to generalize words across different speakers, Houston and Jusczyk (2000) familiarized infants with isolated words (learning materials) spoken by one speaker and then tested them with passages (test materials) containing those words spoken by another speaker. They discovered that at 7.5 months, infants paid more attention to test materials containing familiar words only when both the learning and test materials were produced by speakers of the same sex. By 10.5 months, the speaker-sex effect was no longer observed, indicating that infants' word-form representations become more abstract with age (for similar findings, see Schmale & Seidl, 2009).

In a study focusing on young children, Ryalls and Pisoni (1997) used a word-recognition task where children aged 3–5 years were asked to identify words from a list by pointing to corresponding pictures. The words were spoken by either a single speaker or multiple speakers. Results indicated that children's word recognition was adversely affected by an increased number of speakers. However, as children aged, their ability to process words from multiple speakers improved. Additionally, when asked to repeat the words, younger children matched the duration of the words more closely than older children and adults, suggesting that they retain more acoustic details in their speech representation. These findings imply that infants and young children are more sensitive to acoustic details in speech, with this sensitivity gradually decreasing as they develop (Creel & Tumlin, 2011).

Furthermore, the ability to move beyond these specific acoustic details to generalize across speakers appears to be directly linked to language ability. Levi et al. (2019) investigated the familiar talker advantage in children with varying language abilities. They found that while all children benefitted from familiarity (i.e., successfully mapping specific acoustic details to linguistic units), only those with higher language scores could generalize this knowledge to recognize words spoken by unfamiliar speakers with the same accent. This suggests that while the ability to use speaker-specific acoustic cues is robust even in children with lower language skills, the ability to abstract these patterns to new speakers can indicate higher language ability.

On the other hand, training with multiple speakers can aid speech learning for both first language (Quam & Creel, 2021) and second language (Zhang et al., 2021) acquisition. In an early study, Lively et al. (1993) trained Japanese listeners to distinguish between English/r/and/l/sounds, using either multiple speakers or a single speaker. Those trained with multiple speakers successfully generalized their learning to new words spoken by new speakers, whereas those trained with a single speaker did not. This suggests that exposure to multiple speakers fosters more robust and abstract linguistic representations, which can facilitate the development of phonetic categories and the generalization of speech perception ability. Rost and McMurray (2009, 2010) further explored this idea by showing that acoustic variability aids infants in developing phonetic categories, such as/b/and/p/. Their studies revealed that infants' phonetic learning could be improved by presenting words produced by multiple speakers, compared to presenting words produced by a single speaker. These findings suggest that speaker variability, irrelevant of contrasting phonetic units, can help young language learners acquire those phonetic units (see also Quam et al., 2017). By exposing learners to a wide range of acoustic variations, multi-speaker training may help them extract the invariant features that define phonetic categories, leading to more successful generalization across speakers and contexts.

## Speaker model modulated by a listener's socio-cognitive traits

Language communication is a primary form of social interaction. Consequently, individual differences in social cognition are often reflected in how people process language. Specifically, a listener's socio-cognitive traits may influence their ability to construct a mental model that accurately captures the features of a specific individual or the general attributes of a demographic group.

This link between social cognition and language processing emerges early in life. Kinzler et al. (2007) demonstrated that infants and young children use acoustic cues (e.g., accent) to form social preferences that guide interaction. They found that 5-month-old infants prefer to look at native-language speakers, 10-month-olds prefer to accept toys from native speakers, and 5-year-olds choose to be friends with children who speak with a native accent rather than a foreign accent. For adults, Dragojevic and Giles (2016) showed that processing fluency (i.e., the ease with which speech is processed) acts as a mechanism for social evaluation. They found that when listeners encountered speech that was difficult to process (e.g., due to an unfamiliar accent), they showed a negative affective reaction. This negative affect, in turn, led listeners to evaluate the speaker more negatively.

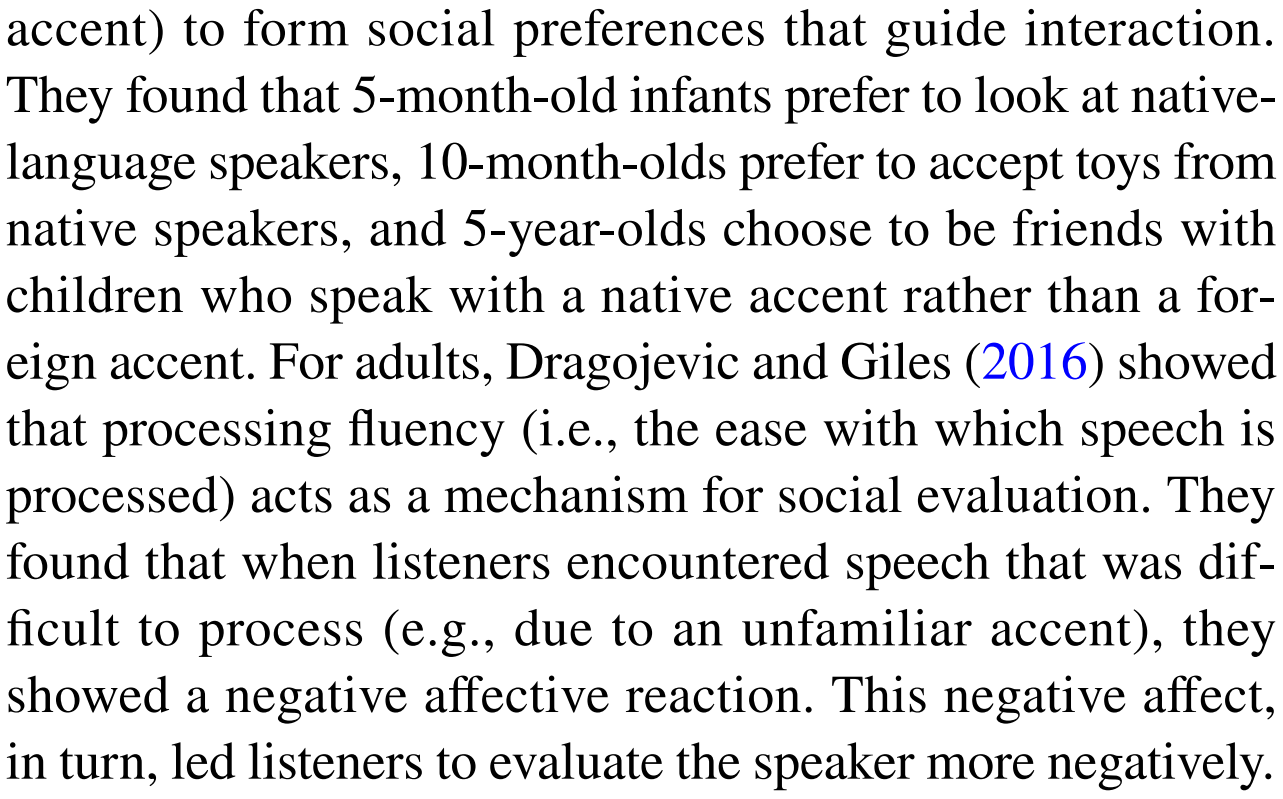

While the ability to extract social identity from voice and speech is a hallmark of typical development, disruptions in voice processing are frequently observed in clinical and neurodiverse populations. Along with phonagnosia (also known as pure voice processing deficit, Hailstone et al., 2010; Van Lancker & Canter, 1982), difficulties in voice processing are observed among populations with schizophrenia, dyslexia, and autism (Stevenage, 2018). Individuals with schizophrenia, particularly those experiencing auditory hallucinations, often struggle to recognize a speaker's identity through voice (Alba-Ferrara et al., 2012; Badcock & Chhabra, 2013; Chhabra et al., 2012). This difficulty is linked to reduced activation in the right STG (Zhang et al., 2008), a region crucial for voice perception (Lattner et al., 2005). Dyslexic individuals generally retain normal facial recognition abilities (Brachacki et al., 1994) but encounter challenges in voice identification (Perea et al., 2014; Perrachione et al., 2011). For autistic individuals, research indicates that challenges in vocal-identity processing often coincide with difficulties in face-identity processing (Boucher et al., 1998), and similar findings are also observed in relation to autistic traits in the general population (Skuk et al., 2019). Individuals with higher autistic traits show reduced activation in the right STS/STG when processing vocal sounds, compared to control individuals (Schelinski et al., 2016).

In the integrative model, deficits in vocal-identity processing may impair the construction and robustness of the speaker model during language comprehension. As the speaker model relies on the listener's ability to extract and process relevant speaker characteristics from the acoustic signal, difficulties in voice processing may lead to a less accurate representation of the speaker. This, in turn, can affect the top-down influence of the speaker model on language comprehension, potentially leading to impairments in the integration of speaker information with linguistic content.

As a direct investigation of this idea, Tesink, Buitelaar et al. (2009a) used fMRI to explore whether autistic individuals differ from non-autistic controls in how they integrate speaker demographics (inferred from speaker voice) with linguistic content during spoken language comprehension.

They found that, compared to control participants, autistic participants showed increased activation in the right inferior frontal gyrus (IFG) for utterances where speaker demographics mismatched the linguistic content, such as "I cannot sleep without my teddy bear in my arms" spoken by an adult speaker. Given their comparable behavioral performance, the authors concluded that it was more difficult for autistic individuals to process speaker properties during language comprehension, and that the heightened IFG activity reflected a cognitive compensation due to increased task demands.

In the general population, speaker effects in language comprehension are influenced by personal traits such as empathy and openness. Using EEG, van den Brink et al. (2012) discovered that individuals with greater empathy showed an increased N400 effect and gamma band oscillatory power when comprehending messages that violated stereotypical expectations associated with the speaker's population. This suggests that more empathetic individuals may have a more detailed or more readily activated demographic speaker model, leading to greater sensitivity to mismatches between the speaker's characteristics and linguistic content. Similarly, Wu and Cai (2026) showed that the magnitude of speaker effects elicited by social stereotypes decreased as a function of the participants' openness trait for both EEG and behavioral measures, as more open-minded people tend to have fewer stereotypical views. Wu et al. (2025) showed that the neural oscillatory response to stereotype-incongruent statements was also modulated by the listener's openness, specifically within the theta frequency band (4–6 Hz). They found that while participants with lower openness scores tended to exhibit increased theta power when encountering incongruent statements (interpreted as reflecting the effortful maintenance of their initial stereotype-based model), those with higher openness scores tended to show a decrease in theta power, suggesting a flexible deployment of attention to updating the speaker model based on the new input. These findings imply that individuals with higher openness not only rely less on fixed demographic stereotypes as priors but also possess greater cognitive flexibility to dynamically update their speaker models when presented with conflicting evidence.

## Future directions: Artificial agents as speakers

Thus far, we have reviewed how language comprehension is influenced by the speaker characteristics ranging from acoustic-episodic memory and shared experience to expectations derived from human demographic categories such as age, gender, and region of origin. However, the rapidly evolving landscape of communication presents a test for the universality of this framework: the emergence of the artificial agent as an interlocutor. As voice-based AI technology transitions from novelty to ambient infrastructure, artificial agents are establishing themselves as a new, synthetic "demographic" group. AI speakers are now ubiquitous in daily life, functioning in various communicative roles, such as virtual assistants (Hoy, 2018), customer service agents (Adam et al., 2021), news anchors (Fitria, 2024), language teachers (Schmidt & Strassner, 2022), navigators (Kun et al., 2007), and even psychotherapists (Fiske et al., 2019). This trend calls for an expansion of research on language comprehension to include artificial agents as a type of speaker and to consider their unique features in studying AI language comprehension.

Research suggests that people often attribute human-like qualities to artificial systems, interacting with them as if they were humans (Nass & Moon, 2000; Reeves & Nass, 1996). This interaction involves applying social norms and behaviors such as politeness (Nass et al., 1999), gender stereotypes (Nass et al., 1997), and reciprocity (Fogg & Nass, 1997). In this sense, artificial agents can be viewed as a particular demographic population of "digital humans," contrasting with the "real human" population. This perspective raises questions about how demographic representations, which are typically based on human social categories, may be adapted or extended to accommodate artificial agents as a unique demographic group.

Studies show that awareness that a speaker is artificial changes the way people interact with them. People tend to control and simplify their language (Amalberti et al., 1993; Kennedy, 1988), exhibit less politeness (Hill et al., 2015), feel less social pressure (Vollmer et al., 2018), and show less desire to establish relationships (Shechtman & Horowitz, 2003) when interacting with artificial agents compared to humans. In the psycholinguistic literature, studies show that people are more likely to reuse lexical expressions (Branigan et al., 2011; Shen & Wang, 2023) previously used by artificial interlocutors than those used by human ones. This tendency for lexical repetition is stronger when interacting with basic artificial systems than advanced ones, possibly in an attempt to enhance understanding with a linguistically limited agent (Branigan et al., 2011; Cai et al., 2021; Pearson et al., 2006).

Despite significant efforts to understand how people interact with artificial agents, limited attention has been paid to how people comprehend their language. Historically, artificial agents were seen as limited in world knowledge (Broussard, 2018) and linguistic capabilities (Kennedy, 1988). However, the development of generative AI has significantly changed the landscape, demonstrating impressive capabilities akin to human creativity (Haase & Hanel, 2023) and language use (Cai et al., 2024). Understanding how AI development influences language comprehension becomes increasingly important, as it may challenge existing assumptions about the limitations of artificial agents.

In one such attempt, Yin et al. (2024) explored whether AI-generated language could make people "feel heard" and whether the "AI label" could influence this feeling. They found that AI-generated messages made participants feel heard to a larger extent than human-generated ones, suggesting that AI was better at detecting emotions in that specific context. However, when participants were told that the messages were from an AI, they felt heard to a lesser extent. This suggests that the "AI identity" affects perceived emotional support in language comprehension. In a direct investigation of how knowing the language is AI-generated influences comprehension, Rao et al. (2025a) used ERPs to test participants' brain responses when encountering semantic and syntactic anomalies perceived as being produced by a large language model (LLM) versus humans. They found that while participants showed overall N400 effects for semantic anomalies and P600 effects for syntactic anomalies, the semantic N400 effects were smaller, and syntactic P600 effects were larger when they were informed that the anomalies were produced by an LLM compared to by a human (see also Rao et al., 2025b, c).

Aside from the influence of artificial agents' non-human nature, a further question is whether this non-human identity interacts with the demographic personas assigned to them. People often attribute traits such as gender, age, and linguistic background to artificial systems. For example, they perceive humanoid artificial agents as male or female based on their appearance (Eyssel & Hegel, 2012) or synthesized voice (Nass et al., 1997). People perceive female agents to be more knowledgeable about dating, using fewer words to explain dating norms compared to male agents (Powers et al., 2005). Similarly, people perceive artificial agents as having a certain age based on their facial features (Powers & Kiesler, 2006) or synthesized voice (Sandygulova & O'Hare, 2015). People are more compliant with requests from agents with a baby face than those with an adult face (Powers & Kiesler, 2006) and prefer a child voice for home companion agents but an adult voice for educational agents (Dou et al., 2021).

These findings align with the idea that people construct an anthropomorphic model of an artificial agent. This anthropomorphic model can be considered a specific type of demographic speaker model, which incorporates expectations about the artificial agent's characteristics and capabilities based on the attributed demographic features (e.g., gender, age). This model can then influence language comprehension in a similar way to the demographic speaker model for human speakers, by biasing the processing of linguistic content and generating expectations about the speaker's knowledge, perspectives, and communicative goals.

However, the extent to which the anthropomorphic model of an artificial agent overlaps with or differs from the demographic speaker model for a human speaker remains an open question. It is possible that people have distinct expectations and biases for artificial agents compared to human speakers, even when they are attributed the same demographic features. For example, people may expect a female artificial agent to have different knowledge and capabilities compared to a female human speaker, due to the perceived differences in their underlying nature and origins. This raises the question of whether findings from human language comprehension can be generalized to AI language comprehension, a research area that remains to be explored.

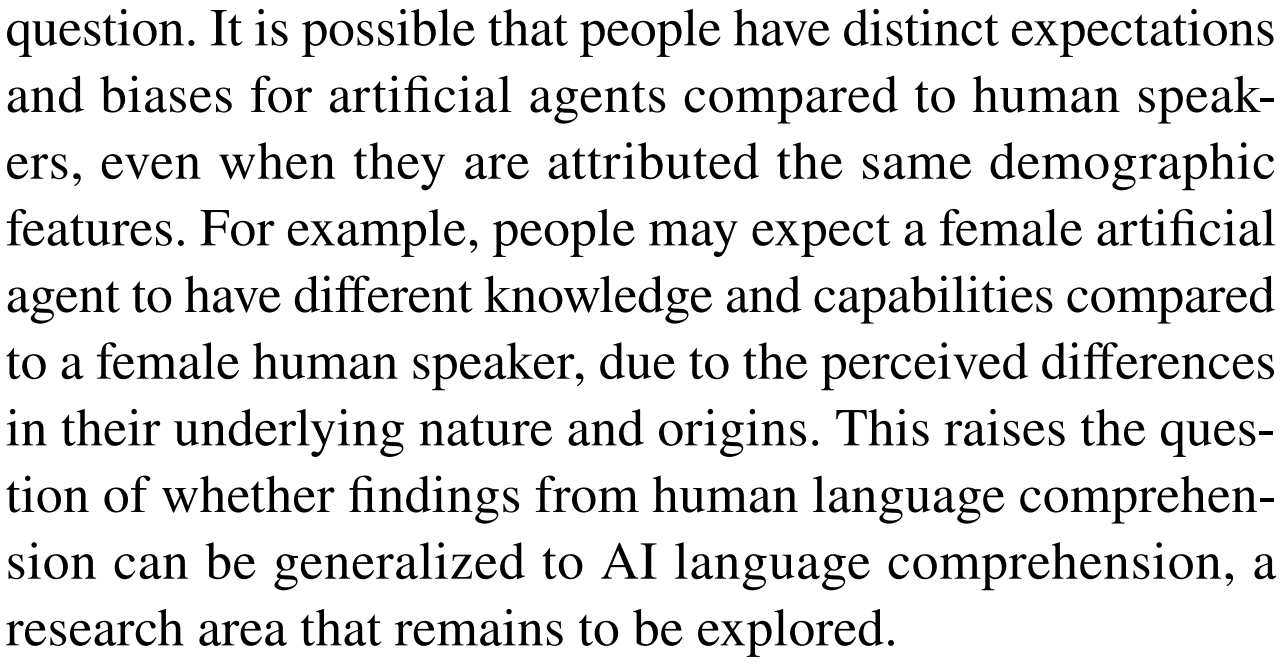

## Conclusion

In this review, we propose an integrative model of language and speaker processing to account for speaker effects in language comprehension. We argue that the influence of a speaker's identity results from the interplay between lower-level acoustic-episodic memory and a higher-level speaker model. We formalize the interaction between language and speaker processing as a bidirectional probabilistic process: prior beliefs about a speaker modulate language comprehension, while the unfolding speech and message continuously updates the speaker model. Within this integrative framework, we define speaker-idiosyncrasy effects and speaker-demographics effects, and show how bottom-up and top-down processes interact at various levels depending on the task and context. We suggest that for studies beyond the psycholinguistic domain, speaker effects can be useful indices of language development and socio-cognitive traits. We encourage future research to explore the applicability of these findings to AI speakers, investigating whether the effects observed with human speakers can be generalized to non-human entities.

**Authors' contributions** Hanlin Wu: Conceptualization, visualization, writing – original draft; Zhenguang G. Cai: Writing – review and editing.

**Funding** This work was supported by the General Research Fund (grant number: 14600220), University Grants Committee, Hong Kong.

**Data availability** Not applicable.

**Code availability** Not applicable.

## Declarations

**Conflicts of interest** The authors have no conflicts of interest to disclose.

**Ethics approval** Not applicable.

**Consent to participate** Not applicable.

**Publisher's Note** Springer Nature remains neutral with regard to jurisdictional claims in published maps and institutional affiliations.